\documentclass[10pt,twocolumn,letterpaper]{article}

\usepackage{cvpr}
\usepackage{times}
\usepackage{epsfig}
\usepackage{graphicx}
\usepackage{amsmath}
\usepackage{amssymb}

\usepackage{url}
\usepackage{epsfig}
\usepackage{mathrsfs}
\usepackage{multirow}
\usepackage{amsmath,bm}
\usepackage{tabu}
\usepackage{rotating}
\usepackage{stfloats}
\usepackage{float}
\usepackage{bbding}
\usepackage{booktabs}

\usepackage[breaklinks=true,bookmarks=false]{hyperref}

\cvprfinalcopy 

\def\cvprPaperID{****} 

\begin{document}

\title{Learning Semantic Concepts and Order for Image and Sentence Matching } 

\author{Yan Huang$^{1,3}$ \hspace{7mm} Qi Wu$^{4}$  \hspace{7mm} Liang Wang$^{1,2,3}$\\
$^1$Center for Research on Intelligent Perception and Computing (CRIPAC),\\
National Laboratory of Pattern Recognition (NLPR)\\
$^2$Center for Excellence in Brain Science and Intelligence Technology (CEBSIT),\\
Institute of Automation, Chinese Academy of Sciences (CASIA)\\
$^3$University of Chinese Academy of Sciences (UCAS)\\
$^4$School of Computer Science, The University of Adelaide\\
{\tt\small \{yhuang, wangliang\}@nlpr.ia.ac.cn \hspace{3mm} qi.wu01@adelaide.edu.au }
}

\maketitle

\begin{abstract}
   Image and sentence matching has made great progress recently, but it
   remains challenging due to the large visual-semantic discrepancy.
   This mainly arises from that the representation of pixel-level image usually lacks of
   high-level semantic information as in its matched sentence.
   In this work, we propose a semantic-enhanced image and sentence matching model,
   which can improve the image representation
   by learning semantic concepts and then organizing them in a correct semantic order.
   Given an image, we first use a multi-regional multi-label CNN
   to predict its semantic concepts, including objects, properties, actions, etc.
   Then, considering that different orders of semantic concepts lead to
   diverse semantic meanings,
   we use a context-gated sentence generation scheme for semantic order learning.
   It simultaneously uses the image global context containing concept relations as reference
   and the groundtruth semantic order in the matched sentence as supervision.
   After obtaining the improved image representation, we learn the sentence representation
   with a conventional LSTM, and then jointly perform image and sentence matching
   and sentence generation for model learning.
   Extensive experiments demonstrate the effectiveness of
   our learned semantic concepts and order,
   by achieving the state-of-the-art results on two public benchmark datasets.

\end{abstract}

\section{Introduction}
The task of image and sentence matching refers to measuring the visual-semantic similarity
between an image and a sentence. It has been widely applied to the application
of image-sentence cross-modal retrieval, \emph{e.g.},
given an image query to find similar sentences, namely image annotation,
and given a sentence query to retrieve matched images, namely text-based image search.

\begin{figure}[t]
\centering
\includegraphics[scale=0.42]{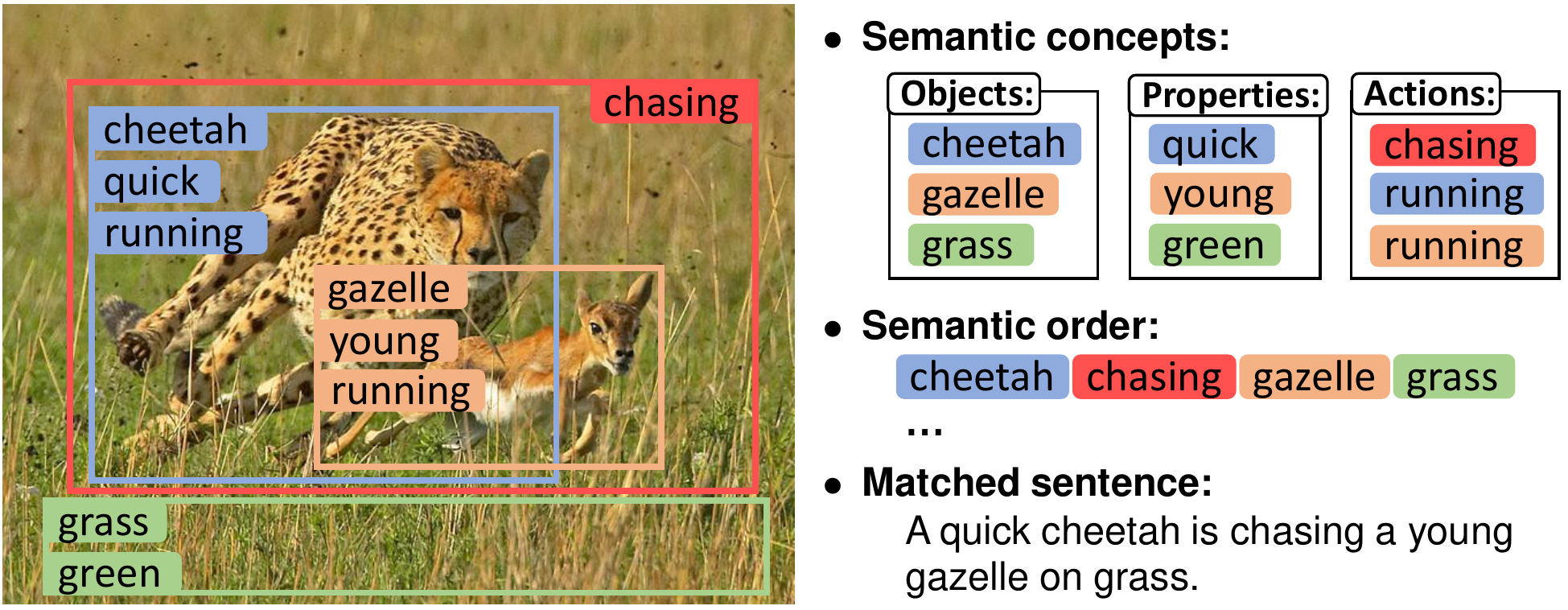}
\caption{Illustration of the semantic concepts and order (best viewed in colors).}
\label{figure:intro}
\end{figure}

Although much progress in this area has been achieved,
it is still nontrivial to accurately measure the similarity between image and sentence,
due to the existing huge visual-semantic discrepancy.
Taking an image and its matched sentence in Figure \ref{figure:intro} for example,
main objects, properties and actions appearing in the image are:
\{\emph{cheetah}, \emph{gazelle}, \emph{grass}\}, \{\emph{quick}, \emph{young}, \emph{green}\}
and \{\emph{chasing}, \emph{running}\}, respectively.
These high-level semantic concepts are the essential content
to be compared with the matched sentence,
but they cannot be easily represented from the pixel-level image.
Most existing methods \cite{huang2016instance,klein2015associating,ma2015multimodal}
jointly represent all the concepts by extracting a global CNN \cite{simonyan2014very} feature vector,
in which the concepts are tangled with each other.
As a result, some primary foreground concepts tend to be dominant,
while other secondary background ones will probably be ignored,
which is not optimal for fine-grained image and sentence matching.
To comprehensively predict all the semantic concepts for the image,
a possible way is to adaptively explore the attribute learning frameworks
\cite{fang2015captions,wu2016value,wei2014cnn}.
But such a method has not been well investigated
in the context of image and sentence matching.

In addition to semantic concepts,
how to correctly organize them, namely semantic order, plays an even more
important role in the visual-semantic discrepancy.
As illustrated in Figure \ref{figure:intro}, given the semantic concepts mentioned above,
if we incorrectly set their semantic order as:
\emph{a quick gazelle is chasing a young cheetah on grass},
then it would have completely different meanings
compared with the image content and matched sentence.
But directly learning the correct semantic order from semantic concepts is very difficult,
since there exist various incorrect orders that semantically make sense.
We could resort to the image global context, since it already indicates the correct semantic order
from the appearing spatial relations among semantic concepts,
\emph{e.g.}, the cheetah is on the left of the gazelle.
But it is unclear how to suitably combine them with the semantic concepts,
and make them directly comparable to the semantic order in the sentence.

Alternatively, we could generate a descriptive sentence from the image
as its representation.
However, the image-based sentence generation itself,
namely image captioning, is also a very challenging problem.
Even those state-of-the-art image captioning methods
cannot always generate very realistic sentences that capture all image details.
The image details are essential to the matching task,
since the global image-sentence similarity is aggregated from local similarities in image details.
Accordingly, these methods cannot achieve very high performance
for image and sentence matching \cite{vinyals2017show,donahue2015long}.

In this work, to bridge the visual-semantic discrepancy between image and sentence,
we propose a semantic-enhanced image and sentence matching model,
which improves the image representation by learning semantic concepts
and then organizing them in a correct semantic order.
To learn the semantic concepts,
we exploit a multi-regional multi-label CNN that can simultaneously
predict multiple concepts in terms of objects, properties, actions, \emph{etc}.
The inputs of this CNN are multiple selectively extracted regions from the image,
which can comprehensively capture all the concepts regardless of whether they are primary foreground ones.
To organize the extracted semantic concepts in a correct semantic order,
we first fuse them with the global context of the image in a gated manner.
The context includes the spatial relations of all the semantic concepts,
which can be used as the reference to facilitate the semantic order learning.
Then we use the groundtruth semantic order in the matched sentence as
the supervision, by forcing the fused image representation to generate the matched sentence.

After enhancing the image representation with both semantic concepts and order,
we learn the sentence representation with a conventional LSTM \cite{hochreiter1997long}.
Then the representations of image and sentence are matched
with a structured objective, which is in conjunction with another objective of
sentence generation for joint model learning.
To demonstrate the effectiveness of the proposed model,
we perform several experiments of image annotation and retrieval
on two publicly available datasets, and achieve the state-of-the-art results.


%

\section{Related Work}


\subsection{Visual-semantic Embedding Based Methods}

Frome \etal \cite{frome2013devise} propose the first visual-semantic embedding framework,
in which ranking loss, CNN \cite{krizhevsky2012imagenet}
and Skip-Gram \cite{mikolov2013efficient} are used as the objective,
image and word encoders, respectively.
Under the similar framework,
Kiros \etal \cite{kiros2014unifying} replace the Skip-Gram with
LSTM \cite{hochreiter1997long}
for sentence representation learning,
Vendrov \etal \cite{vendrov2015order} use a new objective
that can preserve the order structure of visual-semantic hierarchy,
and Wang \etal \cite{wang2015learning} additionally consider within-view constraints
to learn structure-preserving representations.

Yan and Mikolajczyk \cite{yan2015deep} associate the image and sentence
using deep canonical correlation analysis as the objective,
where the matched image-sentence pairs have high correlation.
Based on the similar framework, Klein \etal \cite{klein2015associating}
use Fisher Vectors (FV) \cite{perronnin2007fisher} to
learn more discriminative representations for sentences,
Lev \etal \cite{lev2015rnn} alternatively use RNN to aggregate FV
and further improve the performance,
and Plummer \etal \cite{plummer2015flickr30k} explore the use of
region-to-phrase correspondences.
In contrast, our proposed model considers to bridge the visual-semantic discrepancy
by learning semantic concepts and order.




\begin{figure*}[t]

\centering
\includegraphics[scale=0.57]{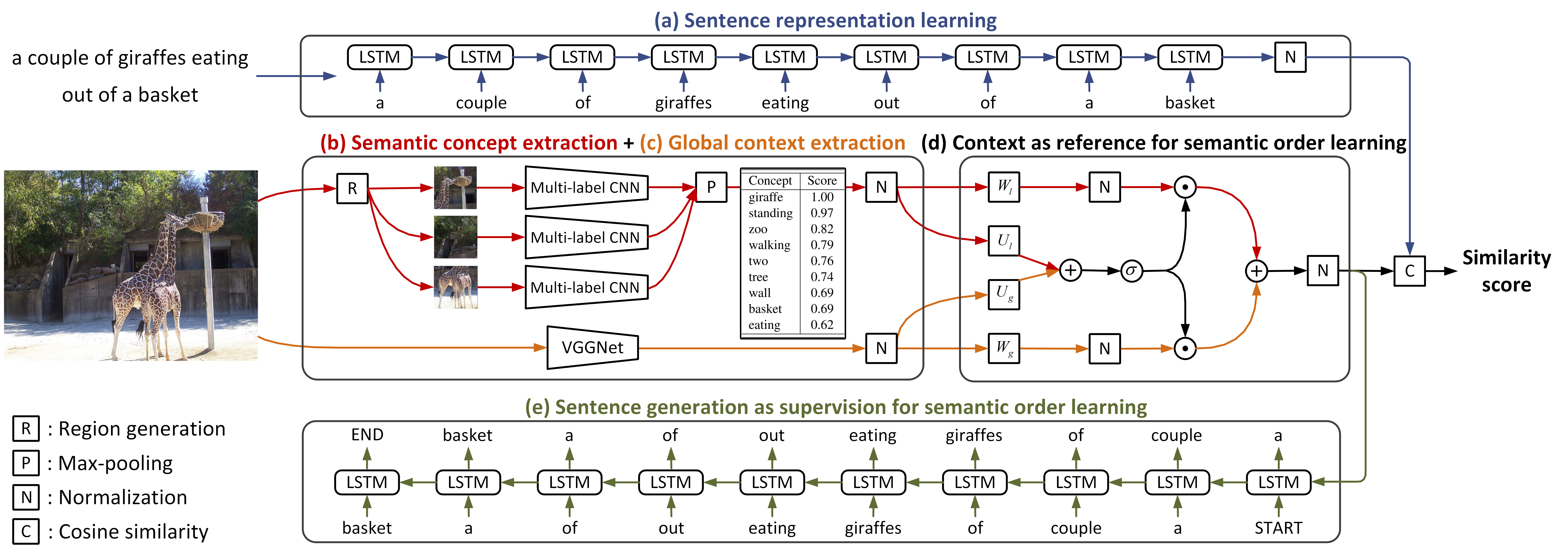}
\caption{The proposed semantic-enhanced image and sentence matching model.}
\label{figure:framework}
\end{figure*}

\subsection{Image Captioning Based Methods}

Chen and Zitnick \cite{chen2015mind} use a multimodal auto-encoder for bidirectional mapping,
and measure the similarity using the cross-modal likelihood and reconstruction error.
Mao \etal \cite{mao2014explain} propose a multimodal RNN model
to generate sentences from images,
in which the perplexity of generating a sentence is used as the similarity.
Donahue \etal \cite{donahue2015long} design a long-term recurrent convolutional network
for image captioning, which can be extended to image and sentence matching as well.
Vinyals \etal \cite{vinyals2017show} develop a neural image captioning generator and
show the effectiveness on the image and sentence matching.
These models are originally designed to
predict grammatically-complete sentences,
so their performance on measuring the image-sentence similarity is not very well.
Different from them, our work focuses on the similarity measurement,
which is especially suitable for the task of image and sentence matching.

\section{Semantic-enhanced Image and Sentence Matching}

In this section, we will detail our proposed semantic-enhanced image and sentence matching
model from the following aspects:
1) sentence representation learning with a conventional LSTM,
2) semantic concept extraction with a multi-regional multi-label CNN,
3) semantic order learning with a context-gated sentence generation scheme,
and 4) model learning with joint image and sentence matching and sentence generation.

\subsection{Sentence Representation Learning}


For a sentence, its included nouns, verbs and adjectives
directly correspond to the visual semantic concepts of object, property and action, respectively,
which are already given.
The semantic order of these semantic-related words
is intrinsically exhibited by the sequential nature of sentence.
To learn the sentence representation that
can capture those semantic-related words
and model their semantic order,
we use a conventional LSTM, similar to \cite{kiros2014unifying,vendrov2015order}.
The LSTM has multiple components for information memorizing and forgetting,
which can well suit the complex properties of semantic concepts and order.
As shown in Figure \ref{figure:framework} (a),
we sequentially feed all the words of the sentence into the LSTM at different timesteps,
and then regard the hidden state at the last timestep as
the desired sentence representation $\textbf{s}\in {\mathbb{R}^{H}}$.

\subsection{Image Semantic Concept Extraction}

For images, their semantic concepts refer to various objects,
properties, actions, \emph{etc}.
The existing datasets do not provide these information at all
but only matched sentences,
so we have to predict them with an additional model.
To learn such a model, we manually build a training dataset
following \cite{fang2015captions,wu2016value}.
In particular, we only keep the nouns, adjectives, verbs and numbers as semantic concepts,
and eliminate all the semantic-irrelevant words from the sentences.
Considering that the size of the concept vocabulary is very large,
we ignore those words that have very low use frequencies.
In addition, we unify the different tenses of verbs,
and the singular and plural forms of nouns
to further reduce the vocabulary size.
Finally, we obtain a vocabulary containing $K$ semantic concepts.
Based on this vocabulary, we can generate the training dataset
by selecting multiple words from sentences as the groundtruth semantic concepts.

Then, the prediction of semantic concepts is
equivalent to a multi-label classification problem.
Many effective models on this problem have been proposed recently \cite{wei2014cnn,wu2016value,wang2016cnn,gong2013deep,wu2015deep},
which mostly learn various CNN-based models as nonlinear
mappings from images to the desired multiple labels.
Similar to \cite{wei2014cnn,wu2016value},
we simply use the VGGNet \cite{simonyan2014very} pre-trained
on the ImageNet dataset \cite{russakovsky2015imagenet}
as our multi-label CNN.
To suit the multi-label classification, we modify the output layer to have $K$ outputs,
each corresponding to the predicted confidence score of a semantic concept.
We then use the sigmoid activation instead of softmax on the outputs,
so that the task of multi-label classification is transformed to multiple tasks of binary classification.
Given an image, its multi-hot representation of
groundtruth semantic concepts is ${{\textbf{y}}_i} \in {\{0,1\}^{K}}$ and
the predicted score vector by the multi-label CNN is
${\widehat {\textbf{y}}_i} \in {[0,1]^{K}}$,
then the model can be learned by optimizing the following objective:
\begin{equation} \label{eqn:e2}
\setlength{\abovedisplayskip}{2pt}
\setlength{\belowdisplayskip}{5pt}
\begin{aligned}
L_{cnn} = {\sum\nolimits_{c = 1}^{K} {\log (1 + {e^{( - {\textbf{y}_{i,c}} {\widehat {\textbf{y}}_{i,c}})}})} }
\end{aligned}
\end{equation}


During testing, considering that the semantic concepts usually appear in
image local regions and vary in size, we perform the concept prediction
in a regional way.
Given a testing image, we first selectively extract $r$ image regions in a similar way
as \cite{wei2014cnn}, and then resize them to square shapes.
As shown in Figure \ref{figure:framework} (b),
by separately feeding these regions into the learned multi-label CNN,
we can obtain a set of predicted confidence score vectors.
Note that the model parameters are shared among all the regions.
We then perform element-wise max-pooling across these score vectors
to obtain a single vector, which includes the desired confidence scores for
all the semantic concepts.



\subsection{Image Semantic Order Learning}
After obtaining the semantic concepts, how to reasonably organize them
in a correct semantic order plays an essential role to the image and sentence matching.
Even though based on the same set of semantic concepts,
combining them in different orders could lead to completely opposite meanings.
For example in Figure \ref{figure:framework} (b), if we organize the extracted
semantic concepts: \emph{giraffes}, \emph{eating} and \emph{basket}
as: \emph{a basket is eating two giraffes}, then its meaning is very
different from the image content.
To learn the semantic order, we propose a context-gated
sentence generation scheme that uses the image global context as reference and the
sentence generation as supervision.

\begin{figure}[t]
\centering
\includegraphics[scale=0.14]{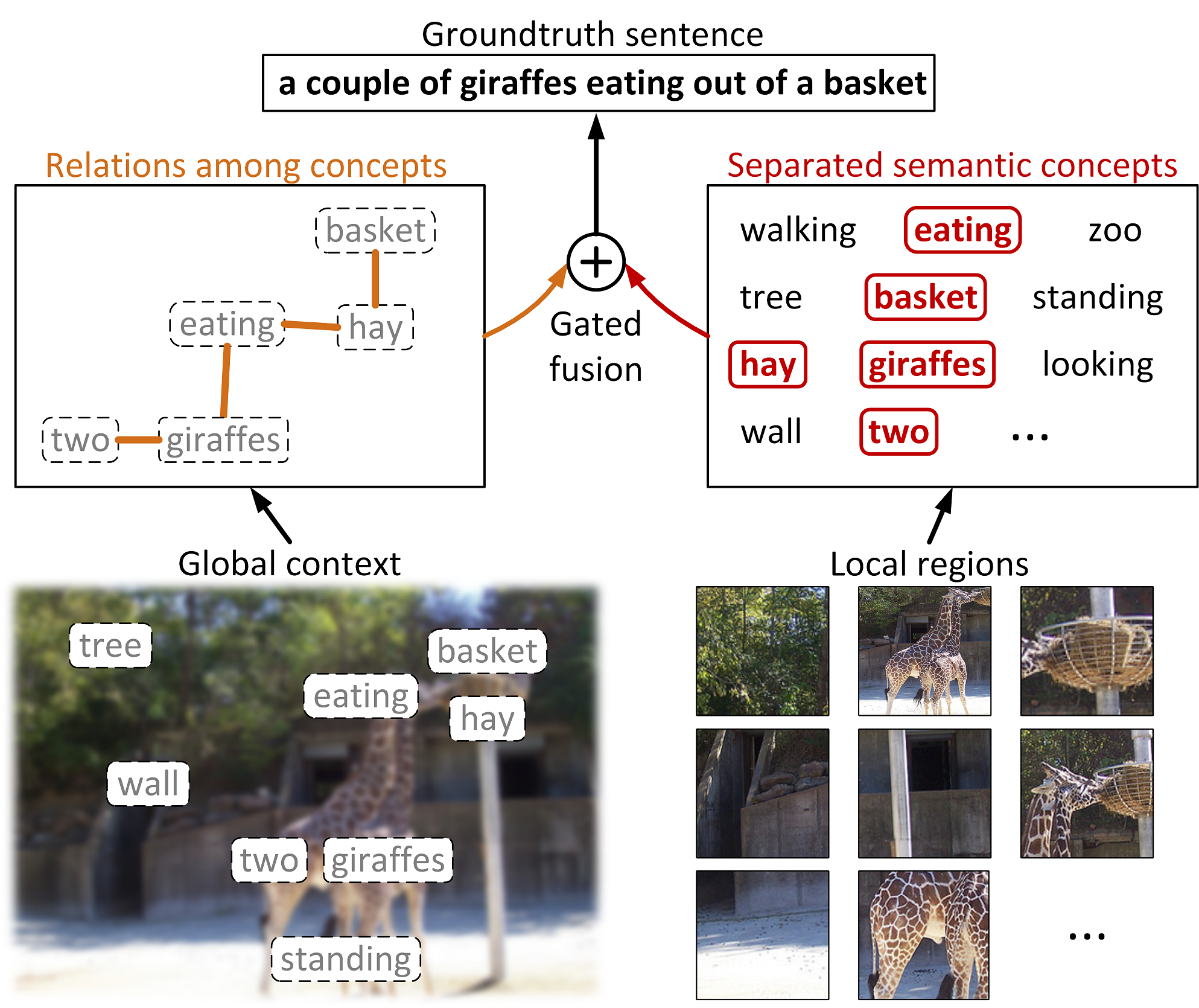}
\caption{Illustration of using the global context as reference for semantic order learning
(best viewed in colors).
}
\label{figure:context}
\end{figure}

\vspace{-3mm}
\subsubsection{Global Context as Reference}
\vspace{-1mm}
It is not easy to learn the semantic order directly from separated semantic concepts,
since the semantic order involves not only the hypernym relations between concepts,
but also the textual entailment among phrases in
high levels of semantic hierarchy \cite{vendrov2015order}.
To deal with this,
we propose to use the image global context as auxiliary reference
for semantic order learning.
As illustrated in Figure \ref{figure:context},
the global context can not only describe all the semantic concepts in a coarse level,
but also indicate their spatial relations with each other,
\emph{e.g.}, two giraffe are standing in the left while the basket is in the top left corner.
When organizing the separated semantic concepts,
our model can refer to the global context
to find their relations and then combine them to facilitate the prediction of semantic order.
In practice, for efficient implementation,
we use a pre-trained VGGNet to process the whole image content,
and then extract the vector in the last fully-connected layer
as the desired global context, as shown in Figure \ref{figure:framework} (c).

To model such a reference procedure,
a simple way is to sum the global context with semantic concepts together.
But considering that the content of different images can be diverse,
thus the relative importance of semantic concepts and context
is not equivalent in most cases.
For those images with complex content,
their global context might be a bit of ambiguous,
so the semantic concepts are more discriminative.
To handle this, we design a gated fusion unit that can selectively balance
the relative importance of semantic concepts and context.
The unit acts as a gate that controls how much information of the
semantic concepts and context contributes to their fused representation.
As illustrated in Figure \ref{figure:framework} (d),
after obtaining the normalized context vector $\textbf{x} \in {\mathbb{R}^{I}}$
and concept score vector $\textbf{p} \in {\mathbb{R}^{K}}$,
their fusion by the gated fusion unit can be
formulated as:
\begin{equation} \label{eqn:e3}
\setlength{\abovedisplayskip}{5pt}
\setlength{\belowdisplayskip}{5pt}
\begin{aligned}
& \widehat {\textbf{p}} = {\left\| {{W_l}\textbf{p}} \right\|_2},{\kern 3pt} \widehat {\textbf{c}} = {\left\| {{W_g}\textbf{x}} \right\|_2}
, {\kern 3pt} \textbf{t} = \sigma ({U_l}\textbf{p} + {U_g}\textbf{x})\\
& \textbf{v} = \textbf{t} \odot \widehat {\textbf{p}} + (\textbf{1} - \textbf{t}) \odot \widehat {\textbf{x}}\\
\end{aligned}
\end{equation}
where ${\left\| \cdot \right\|_2}$ denotes the $l_2$-normalization,
and $\textbf{v}\in {\mathbb{R}^{H}}$ is the fused representation of semantic concepts and global context.
The use of sigmoid function $\sigma$ is to rescale each element in the gate vector $\textbf{t}\in {\mathbb{R}^{H}}$ to $[0,1]$,
so that $\textbf{v}$ becomes an element-wise weighted sum of
$\textbf{p}$ and $\textbf{x}$.

\vspace{-3mm}
\subsubsection{Sentence Generation as Supervision} \label{sent:order}
\vspace{-1mm}
To learn the semantic order based on the fused representation,
a straightforward approach is to directly generate a sentence from it,
similar to image captioning \cite{wu2016value}.
However, such an approach is infeasible resulting from the following problem.
Although the current image captioning methods can
generate semantically meaningful sentences,
the accuracy of their generated sentences on capturing image details is not very high.
And even a little error in the sentences can be amplified
and further affect the measurement of similarity,
since the generated sentences are highly semantic and the similarity is computed in a fine-grained level.
Accordingly, even the state-of-the-art image captioning models
\cite{vinyals2017show,donahue2015long,mao2014explain}
cannot perform very well on the image and sentence matching task.
We also implement a similar model (as ``ctx + sen'') in Section \ref{sent:ablation},
but find it only achieves inferior results.


In fact, it is unnecessary for the image and sentence matching task
to generate a grammatically-complete sentence.
We can alternatively regard the fused context and concepts as the image representation,
and supervise it using the groundtruth semantic order in the matched sentence
during the sentence generation.
As shown in Figure \ref{figure:framework} (e),
we feed the image representation into the initial hidden state
of a generative LSTM, and ask it to be capable of generating the matched sentence.
During the cross-word and cross-phrase generations,
the image representation can thus learn
the hypernym relations between words and textual entailment among phrases
as the semantic order.

Given a sentence
$\left\{ {{{\textbf{w}}_j}}| {{\textbf{w}}_j} \in {\{0,1\}^{G}}  \right\}_{j=1,\cdots,J}$,
where each word ${{\textbf{w}}_j}$ is represented as an one-hot vector,
$J$ is the length of the sentence, and $G$ is the size of word dictionary,
we can formulate the sentence generation as follows:
\begin{equation}\label{eqn:e4}
\setlength{\abovedisplayskip}{5pt}
\setlength{\belowdisplayskip}{5pt}
\begin{aligned}
&{\textbf{i}_t} = \sigma ({W_{\textbf{w}\textbf{i}}}(F{\textbf{w}_t}) + {W_{\textbf{h}\textbf{i}}}{\textbf{h}_{t - 1}} + {\textbf{b}_\textbf{i}}), \\
&{\textbf{f}_t} = \sigma ({{W}_{\textbf{w}\textbf{f}}}(F{\textbf{w}_t}) + {W_{\textbf{h}\textbf{f}}}{\textbf{h}_{t - 1}} + {\textbf{b}_\textbf{f}}),\\
&{\textbf{o}_t} = \sigma ({W_{\textbf{w}\textbf{o}}}(F{\textbf{w}_t}) + {W_{\textbf{h}\textbf{o}}}{\textbf{h}_{t - 1}} + {\textbf{b}_\textbf{o}}),\\
&\widehat {\textbf{c}_t} = \tanh ({W_{\textbf{w}\textbf{c}}}(F{\textbf{w}_t}) + {W_{\textbf{h}\textbf{c}}}{\textbf{h}_{t - 1}} + {\textbf{b}_\textbf{c}}),\\
&{\textbf{c}_t} = {\textbf{f}_t}\odot{\textbf{c}_{t - 1}} + {\textbf{i}_t}\odot\widehat {\textbf{c}_t},  {\kern 3pt}  {\textbf{h}_t} = {\textbf{o}_t}\odot\tanh ({\textbf{c}_t}),\\
&{\textbf{q}_{t}} = \emph{softmax}({F^T}{\textbf{h}_{t}} + {\textbf{b}_p})
, {\kern 3pt} e = \arg \max ({\textbf{w}_{t}}), \\
&P(\textbf{w}_{t}|\textbf{w}_{t-1}, \textbf{w}_{t-2},\cdots,\textbf{w}_0,\textbf{x},\textbf{p}) = {\textbf{q}_{t,e}}
\end{aligned}
\end{equation}
where ${\textbf{c}_t}$, ${\textbf{h}_t}$, $\textbf{i}_{t}$,
$\textbf{f}_{t}$ and $\textbf{o}_{t}$ are memory state, hidden state,
input gate, forget gate and output gate, respectively,
$e$ is the index of ${\textbf{w}_{t}}$ in the word vocabulary,
and $F\in {\mathbb{R}^{D \times G}}$ is a word embedding matrix.
During the sentence generation,
since all the words are predicted in a chain manner,
the probability $P$ of current predicted word
is conditioned on all its previous words,
as well as the input semantic concepts ${\textbf{p}}$ and context ${\textbf{x}}$ at the initial timestep.

\begin{table*}[t] \small
\addtolength{\tabcolsep}{-2pt}
\centering
\caption{The experimental settings of ablation models.}
\begin{tabular}{l|cc|cc|cc|ccc|cc}
\hline
\hline

       & 1-crop & 10-crop  & context  & concept  & sum  & gate & sentence  & generation  & sampling & shared & non-shared\\
\hline

ctx (1-crop)          & ${\surd}$ & &${\surd}$ & & & & & & & & \\
ctx                     & &${\surd}$ &${\surd}$ & & & & & & & & \\
\hline
ctx + sen     & &${\surd}$ &${\surd}$ & & & &${\surd}$ & & & &${\surd}$ \\
ctx + gen (S)     & &${\surd}$ &${\surd}$ & & & & &${\surd}$ &${\surd}$ & &${\surd}$ \\
ctx + gen (E)     & &${\surd}$ &${\surd}$ & & & & &${\surd}$ & &${\surd}$ & \\
ctx + gen                 & &${\surd}$ &${\surd}$ & & & & &${\surd}$ & & &${\surd}$ \\
\hline
cnp                   & & & &${\surd}$ & & & & & & & \\
cnp + gen          & & & &${\surd}$ & & & &${\surd}$ & & & \\
cnp + ctx (C)     & &${\surd}$ &${\surd}$ &${\surd}$ &${\surd}$ & & & & & & \\
cnp + ctx        & &${\surd}$ &${\surd}$ &${\surd}$ & &${\surd}$ & & & & & \\
\hline
cnp + ctx + gen     & &${\surd}$ &${\surd}$ &${\surd}$ & &${\surd}$ & &${\surd}$ & & &${\surd}$ \\

\hline
\hline
\end{tabular}

\label{table:ablation}
\end{table*}

\subsection{Joint Matching and Generation}

During the model learning, to jointly perform image and sentence matching and sentence generation,
we need to minimize the following combined objectives:
\begin{equation} \label{eqn:e6}
\setlength{\abovedisplayskip}{5pt}
\setlength{\belowdisplayskip}{5pt}
L = L_{mat} + \lambda \times L_{gen}
\end{equation}
where $\lambda$ is a tuning parameter for balancing.

The $L_{mat}$ is a structured objective
that encourages the cosine similarity scores of matched images and sentences to be larger
than those of mismatched ones:
\begin{equation*}
\setlength{\abovedisplayskip}{5pt}
\setlength{\belowdisplayskip}{5pt}
\begin{aligned}
\sum\nolimits_{ik}{{\max \left\{ {0,m - s_{ii}+ s_{ik}} \right\}}}  + {{\max \left\{ {0,m - s_{ii} + s_{ki}} \right\}} }
\end{aligned}
\end{equation*}
where $m$ is a margin parameter,
$s_{ii}$ is the score of matched $i$-th image and $i$-th sentence,
$s_{ik}$ is the score of mismatched $i$-th image and $k$-th sentence,
and vice-versa with $s_{ki}$.
We empirically set the total number of mismatched pairs for each matched pair
as 128 in our experiments.

The $L_{gen}$ is the negative conditional log-likelihood of
the matched sentence given the semantic concepts $\textbf{p}$ and context $\textbf{x}$:
\begin{equation*}
\setlength{\abovedisplayskip}{5pt}
\setlength{\belowdisplayskip}{5pt}
 - \sum\nolimits_t {\log P(\textbf{w}_{t}|\textbf{w}_{t-1}, \textbf{w}_{t-2},\cdots,\textbf{w}_0,\textbf{x},\textbf{p})}
\end{equation*}
where the detailed formulation of probability $P$ is shown in Equation \ref{eqn:e4}.
Note that we use the predicted semantic concepts rather than
groundtruth ones in our experiments.

All modules of our model excepting for the multi-regional multi-label CNN
can constitute a whole deep network,
which can be jointly trained in an end-to-end manner from raw image and sentence to their similarity score.
It should be noted that we do not need to generate the sentence during testing.
We only have to compute the image representation $\textbf{v}$
from $\textbf{x}$ and $\textbf{p}$,
and then compare it with the sentence representation $\textbf{s}$
to obtain their cosine similarity score.

\section{Experimental Results}
To demonstrate the effectiveness of the proposed model,
we perform several experiments in terms of image annotation
and retrieval on two publicly available datasets.

\subsection{Datasets and Protocols}

The two evaluation datasets and their experimental protocols are described as follows.
1) {Flickr30k} \cite{young2014image}
consists of 31783 images collected from the Flickr website.
Each image is accompanied with 5 human annotated sentences.
We use the public training, validation and testing splits \cite{kiros2014unifying}, which
contain 28000, 1000 and 1000 images, respectively.
2) {MSCOCO} \cite{lin2014microsoft} consists of 82783
training and 40504 validation images, each of which is associated with 5 sentences.
We use the public training, validation and testing splits \cite{kiros2014unifying},
with 82783, 4000 and 1000 (or 5000) images, respectively.
When using 1000 images for testing, we perform 5-fold cross-validation
and report the averaged results.

\begin{table*}[t] \small
\addtolength{\tabcolsep}{-1pt}
\centering
\caption{Comparison results of image annotation and retrieval by ablation models on the Flickr30k and MSCOCO (1000 testing) datasets. }
\begin{tabular}{l|ccc|ccc|c|ccc|ccc|c}
\hline
\hline
\multirow{3}{0.7cm}{Method}  &  \multicolumn{7}{c|}{Flickr30k dataset} &  \multicolumn{7}{c}{MSCOCO dataset}\\
\cline{2-15}
        &  \multicolumn{3}{c|}{Image Annotation}  &  \multicolumn{3}{c|}{Image Retrieval} & \multirow{2}{0.5cm}{{mR}}
&  \multicolumn{3}{c|}{Image Annotation}  &  \multicolumn{3}{c|}{Image Retrieval} & \multirow{2}{0.5cm}{{mR}}  \\
\cline{2-7}
\cline{9-14}
     & R$@$1 & R$@$5  & R$@$10   & R$@$1 & R$@$5  & R$@$10  &
&R$@$1 & R$@$5  & R$@$10   & R$@$1 & R$@$5  & R$@$10  &    \\
\hline

\hspace{0mm} ctx (1-crop)             &29.8 &58.4 &70.5 &22.0 &47.9 &59.3 &48.0  &43.3 &75.7 &85.8 &31.0 &66.7 &79.9 &63.8\\
\hspace{0mm} ctx                      &33.8 &63.7 &75.9 &26.3 &55.4 &67.6 &53.8  &44.7 &78.2 &88.3 &37.0 &73.2 &85.7 &67.9\\
\hline
\hspace{0mm} ctx + sen               &22.8 &48.6 &60.8 &19.1 &46.0 &59.7 &42.8  &39.2 &73.3 &85.5 &32.4 &70.1 &83.7 &64.0\\
\hspace{0mm} ctx + gen (S)           &34.4 &64.5 &77.0 &27.1 &56.3 &68.3 &54.6  &45.7 &78.7 &88.7 &37.3 &73.8 &85.8 &68.4\\
\hspace{0mm} ctx + gen (E)           &35.5 &63.8 &75.9 &27.4 &55.9 &67.6 &54.3  &46.9 &78.8 &89.2 &37.3 &73.9 &85.9 &68.7\\
\hspace{0mm} ctx + gen                &35.6 &66.3 &76.9 &27.9 &56.8 &68.2 &55.3  &46.9 &79.2 &89.3 &37.9 &74.0 &85.9 &68.9\\
\hline
\hspace{0mm} cnp                     &30.9 &60.9 &72.4 &23.1 &52.5 &64.8 &50.8  &59.5 &86.9 &93.6 &48.5 &81.4 &90.9 &76.8\\
\hspace{0mm} cnp + gen               &31.5 &61.7 &74.5 &25.0 &53.4 &64.9 &51.8  &62.6 &89.0 &94.7 &50.6 &82.4 &91.2 &78.4\\
\hspace{0mm} cnp + ctx (C)          &39.9 &71.2 &81.3 &31.4 &61.7 &72.8 &59.7  &62.8 &89.2 &95.5 &53.2 &85.1 &93.0 &79.8\\
\hspace{0mm} cnp + ctx              &42.4 &72.9 &81.5 &32.4 &63.5 &73.9 &61.1  &65.3 &90.0 &96.0 &54.2 &85.9 &93.5 &80.8\\
\hline
\hspace{0mm} cnp + ctx + gen          &\bf{44.2} &\bf{74.1} &\bf{83.6} &\bf{32.8} &\bf{64.3} &\bf{74.9} &\bf{62.3}   &\bf{66.4} &\bf{91.3} &\bf{96.6} &\bf{55.5} &\bf{86.5} &\bf{93.7} &\bf{81.8}\\

\hline
\hline
\end{tabular}
\label{table:ablation_result}
\end{table*}

\begin{table*}[t] \small
\addtolength{\tabcolsep}{-1pt}
\centering
\caption{Comparison results of image annotation and retrieval on the MSCOCO (1000 testing) dataset. }
\begin{tabular}{l|ccc|ccc|c|ccc|ccc|c}
\hline
\hline
\multirow{3}{0.7cm}{$\lambda$}  &  \multicolumn{7}{c|}{Flickr30k dataset} &  \multicolumn{7}{c}{MSCOCO dataset}\\
\cline{2-15}
        &  \multicolumn{3}{c|}{Image Annotation}  &  \multicolumn{3}{c|}{Image Retrieval} & \multirow{2}{0.5cm}{{mR}}
&  \multicolumn{3}{c|}{Image Annotation}  &  \multicolumn{3}{c|}{Image Retrieval} & \multirow{2}{0.5cm}{{mR}}  \\
\cline{2-7}
\cline{9-14}
     & R$@$1 & R$@$5  & R$@$10   & R$@$1 & R$@$5  & R$@$10  &
&R$@$1 & R$@$5  & R$@$10   & R$@$1 & R$@$5  & R$@$10  &    \\
\hline

0       &42.4 &72.9 &81.5 &32.4 &63.5 &73.9  &61.1   &65.3 &90.0 &96.0 &54.2 &85.9 &93.5  &80.8\\
0.01    &43.1 &72.8 &83.5 &\textbf{32.8} &63.2 &73.6  &61.5   &66.3 &91.2 &96.5 &55.4 &86.5 &93.7  &81.6\\
1       &\bf{44.2} &\bf{74.1} &\bf{83.6} &\bf{32.8} &\bf{64.3} &\bf{74.9}  &\bf{62.3}   &\bf{66.6} &\bf{91.8} &\bf{96.6} &\bf{55.5} &\bf{86.6} &\bf{93.8}  &\bf{81.8}\\
100     &42.3 &73.8 &83.1 &32.5 &63.3 &74.0  &61.5   &65.0 &90.5 &96.1 &54.9 &86.3 &93.7  &81.1\\

\hline
\hline
\end{tabular}
\label{table:lambda}
\end{table*}

\subsection{Implementation Details} \label{sent:details}

The commonly used evaluation criterions for image annotation and retrieval
are ``$\rm R@1$'', ``$\rm R@5$'' and ``$\rm R@10$'',
\emph{i.e.}, recall rates at the top 1, 5 and 10 results.
We also compute an additional criterion ``mR'' by averaging
all the 6 recall rates, to evaluate the overall performance
for both image annotation and retrieval.

For images, the dimension of global context
is $I$=$4096$ for VGGNet \cite{simonyan2014very}
or $I$=$1000$ for ResNet \cite{he2016deep}.
We perform 10-cropping \cite{klein2015associating} from the images
and then separately feed the cropped regions into the network.
The final global context is averaged over 10 regions.
For sentences, the dimension of embedded word is $D$=$300$.
We set the max length for all the sentences as 50,
\emph{i.e.}, the number of words $J$=$50$,
and use zero-padding when a sentence is not long enough.
Other parameters are empirically set as follows:
$H$=$1024$, $K$=$256$, $\lambda$=$1$, $r$=$50$ and $m$=$0.2$.

To systematically evaluate the contributions of different
model components, we design various ablation models
as shown in Table \ref{table:ablation}.
The variable model components are explained as follows:
1) ``1-crop'' and ``10-crop'' refer to cropping 1 or 10 regions from
images, respectively, when extracting the global context.
2) ``concept'' and ``context'' denote using semantic concepts and global context, respectively.
3) ``sum'' and ``gate'' are two different ways that combine semantic concepts
and context via feature summation and gated fusion unit, respectively.
4) ``sentence'', ``generation'' and ``sampling'' are three different ways to learn
the semantic order, in which ``sentence'' uses the state-of-the-art
image captioning method \cite{vinyals2017show} to generate sentences from
images and then regard the sentences as the image representations,
``generation'' uses the sentence generation as supervision
as described in Section \ref{sent:order}, and ``sampling'' additionally uses
the scheduled sampling \cite{bengio2015scheduled}.
5) ``share'' and ``non-shared'' indicate whether the parameters of
two word embedding matrices for sentence representation learning
and sentence generation are shared or not.

\subsection{Evaluation of Ablation Models} \label{sent:ablation}

The results of the ablation models on the Flickr30k and MSCOCO datasets
are shown in Table \ref{table:ablation_result},
from which we can obtain the following conclusions.
1) Cropping 10 image regions (as ``ctx'') can achieve much robust global
context features than cropping only 1 region (as ``ctx (1-crop)'').
2) Directly using the pre-generated sentences as image representations
(as ``ctx + sen'') cannot improve the performance,
since the generated sentences might not accurately
include the image details.
3) Using the sentence generation as supervision
for semantic order learning (as ``ctx + gen'') is very effective.
But additionally performing the scheduled sampling (as ``ctx + gen (S)'') cannot
further improve the performance. It is probably because the groundtruth semantic order
is degenerated during sampling, accordingly the model cannot learn it well.
4) Using a shared word embedding matrix (as ``ctx + gen (E)'') cannot improve
the performance, which might result from that learning a unified
matrix for two tasks is difficult.
5) Only using the semantic concepts (as ``cnp'') can already achieve good performance,
especially when the training data are sufficient on the MSCOCO dataset.
6) Simply summing the concept and context (as ``cnp + ctx (C)'')
can further improve the result,
because the context contains the spatial relations of concepts which are very useful.
7) Using the proposed gated fusion unit (as ``cnp + ctx'') performs better, due to
the effective importance balancing scheme.
8) The best performance is achieve by the ``cnp + ctx + gen'',
which combines the 10-cropped extracted context
with semantic concepts via the gated fusion unit,
and exploits the sentence generation for semantic order learning.
Without using either semantic concepts (as ``ctx + gen'') or context (as ``cnp + gen''),
the performance drops heavily.
In the follow experiments, we regard the ``cnp + ctx + gen'' as our default model.

In addition, we test the balancing parameter $\lambda$
in Equation \ref {eqn:e6}, by varying it from 0 to 100.
The corresponding results are presented in Table \ref{table:lambda},
we can find that when $\lambda$=$1$, the model can achieve its best performance.
It indicates that the generation objective plays an equally important role as the
matching objective.

\begin{table*}[t] \small
\addtolength{\tabcolsep}{-1.5pt}
\centering
\caption{Comparison results of image annotation and retrieval on the Flickr30k and MSCOCO (1000 testing) datasets. }
\begin{tabular}{l|ccc|ccc|c|ccc|ccc|c}
\hline
\hline
\multirow{3}{0.7cm}{Method}  &  \multicolumn{7}{c|}{Flickr30k dataset} &  \multicolumn{7}{c}{MSCOCO dataset}\\
\cline{2-15}
        &  \multicolumn{3}{c|}{Image Annotation}  &  \multicolumn{3}{c|}{Image Retrieval} & \multirow{2}{0.5cm}{{mR}}
&  \multicolumn{3}{c|}{Image Annotation}  &  \multicolumn{3}{c|}{Image Retrieval} & \multirow{2}{0.5cm}{{mR}}  \\
\cline{2-7}
\cline{9-14}
     & R$@$1 & R$@$5  & R$@$10   & R$@$1 & R$@$5  & R$@$10  &
&R$@$1 & R$@$5  & R$@$10   & R$@$1 & R$@$5  & R$@$10  &    \\
\hline

m-RNN \cite{mao2014explain}                &35.4 &63.8 &73.7  &22.8 &50.7 &63.1 & 51.6   &41.0 &73.0 &83.5 &29.0 &42.2 &77.0  & 57.6\\
FV \cite{klein2015associating}             &35.0 &62.0 &73.8 &25.0 &52.7 &66.0  & 52.4   &39.4 &67.9 &80.9 &25.1 &59.8 &76.6  & 58.3\\
DVSA \cite{karpathy2014vsa}                &22.2 &48.2 &61.4  &15.2 &37.7 &50.5 & 39.2   &38.4 &69.9 &80.5 &27.4 &60.2 &74.8  & 58.5\\
MNLM \cite{kiros2014unifying}              &23.0 &50.7 &62.9  &16.8 &42.0 &56.5  &42.0   &43.4 &75.7 &85.8 &31.0 &66.7 &79.9  & 63.8\\
m-CNN \cite{ma2015multimodal}              &33.6 &{64.1} &{74.9}  &{26.2} &{56.3} &{69.6} & 54.1   &42.8 &73.1 &84.1 &32.6 &{68.6} &{82.8} & 64.0\\
RNN+FV \cite{lev2015rnn}                   &34.7 &62.7 &72.6  &{26.2} &55.1 &69.2   & 53.4   &40.8 &71.9 &83.2 &29.6 &64.8 &80.5 & 61.8\\
OEM \cite{vendrov2015order}                &-  &-  &-  &-  &-  &-  &-  &46.7 &78.6 &88.9 &37.9 &73.7 &85.9 &68.6\\
VQA \cite{lin2016leveraging}               &33.9 &62.5  &74.5  &24.9 &52.6 &64.8 & 52.2   &50.5 &80.1 &89.7 &37.0 &70.9 &82.9 & 68.5\\
RTP  \cite{plummer2015flickr30k}           &{37.4} &63.1 &74.3 &26.0 &{56.0} &{69.3} & 54.3    &-  &-  &-  &-  &-  &-  &-  \\
DSPE \cite{wang2015learning}               &40.3 &68.9 &79.9   &29.7 &60.1 &72.1 & 58.5   &50.1 &79.7 &89.2 &39.6 &75.2 &86.9 & 70.1\\
sm-LSTM \cite{huang2016instance}           &{42.5} &{71.9} &{81.5} &{30.2} &{60.4} &{72.3}&{59.8}   &{53.2} &{83.1} &{91.5} &{40.7} &{75.8} &{87.4} & {72.0}\\
2WayNet \cite{Eisenschtat_2017_CVPR}       &\bf{49.8} &67.5 &-   &\bf{36.0} &55.6 &-  & -   &55.8 &75.2 &-  &39.7 &63.3 &- & -\\
DAN \cite{nam2016dual}                     &41.4 &73.5 &82.5  &31.8 &61.7 &72.5   & 60.6   &-  &-  &-  &-  &-  &-  &-  \\
VSE++  \cite{faghri2017vse++}              &41.3 &69.0  &77.9    &31.4 &59.7 &71.2  & 58.4  &57.2 &85.1  &93.3   &45.9 &78.9 &89.1 & 74.6\\
\textbf{Ours}                   &44.2 &\bf{74.1} &\bf{83.6}  &32.8&\bf{64.3} &\bf{74.9} &\bf{62.3}   &\bf{66.6} &\bf{91.8} &\bf{96.6} &\bf{55.5} &\bf{86.6} &\bf{93.8} &\bf{81.8}\\

\hline
RRF (Res) \cite{Learning2017Liu}       &47.6 &77.4 &87.1  &35.4 &68.3 &79.9 & 66.0  &56.4 &85.3 &91.5 &43.9 &78.1 &88.6 &73.9\\
DAN (Res) \cite{nam2016dual}       &55.0 &81.8 &89.0 &39.4 &69.2 &79.1   & 68.9  &-  &-  &-  &-  &-  &-  &-  \\
VSE++ (Res) \cite{faghri2017vse++}      &52.9 &79.1  &87.2     &39.6 &69.6 &79.5  & 68.0   &64.6 &89.1  &95.7  &52.0 &83.1 &92.0  & 79.4\\
\textbf{Ours (Res)}                 &\bf{55.5} &\bf{82.0} &\bf{89.3}  &\bf{41.1} &\bf{70.5} &\bf{80.1}  &\bf{69.7}   &\bf{69.9} &\bf{92.9} &\bf{97.5} &\bf{56.7} &\bf{87.5} &\bf{94.8} &\bf{83.2}\\

\hline
\hline
\end{tabular}
\label{table:state}
\end{table*}

\subsection{Comparison with State-of-the-art Methods}

We compare our proposed model with several recent state-of-the-art models
on the Flickr30k and MSCOCO datasets in Table \ref{table:state}.
The methods marked by ``(Res)'' use the 152-layer ResNet \cite{he2016deep} for context extraction,
while the rest ones use the default 19-layer VGGNet \cite{simonyan2014very}.

Using either VGGNet or ResNet on the MSCOCO dataset,
our proposed model outperforms the current state-of-the-art
models by a large margin on all 7 evaluation criterions.
It demonstrates that learning semantic concepts and order
for image representations is very effective.
When using VGGNet on the Flickr30k dataset,
our model gets lower performance than 2WayNet on the R@1 evaluation criterion,
but obtains much better overall performance on the rest evaluation criterions.
When using ResNet on the Flickr30k dataset, our model is able to achieve the best result.
Note that our model obtains much larger improvements on the MSCOCO dataset than Flickr30k.
It is because the MSCOCO dataset has more training data,
so that our model can be better fitted
to predict more accurate image-sentence similarities.

The above experiments on the MSCOCO dataset follow
the first protocol \cite{karpathy2014vsa},
which uses 1000 images and their associated sentences for testing.
We also test the second protocol that uses all the 5000 images and their sentences
for testing,
and present the comparison results in Table \ref{table:coco5k}.
From the table we can observe that
the overall results by all the methods are lower
than the first protocol.
It probably results from that the target set is much larger
so there exist more distracters for a given query.
Among all the models, the proposed model still achieves the best performance, which again
demonstrates its effectiveness.
Note that our model has much larger improvements using VGGNet than ResNet, which results from
that ``Ours (Res)'' only uses the ResNet for extracting global context
but not semantic concepts.

\begin{table}[t] \small
\addtolength{\tabcolsep}{-3pt}
\centering
\caption{Comparison results of image annotation and retrieval on the MSCOCO (5000 testing) dataset. }
\begin{tabular}{l|ccc|ccc|c}
\hline
\hline
\multirow{2}{0.7cm}{Method}     &  \multicolumn{3}{c|}{Image Annotation}  &  \multicolumn{3}{c|}{Image Retrieval} & \multirow{2}{0.5cm}{{mR}}  \\
\cline{2-7}
     & R$@$1 & R$@$5  & R$@$10  & R$@$1 & R$@$5  & R$@$10  &    \\
\hline

DVSA \cite{karpathy2014vsa}               &11.8 &32.5 &45.4  &8.9 &24.9 &36.3  &26.6 \\
FV\cite{klein2015associating}  &17.3 &39.0 &50.2  &10.8 &28.3 &40.1  &31.0\\
VQA \cite{lin2016leveraging}              &23.5 &50.7 &63.6   &16.7 &40.5 &53.8   & 41.5\\
OEM \cite{vendrov2015order}               &23.3 &50.5 &65.0 &18.0 &43.6   &57.6  &43.0\\
VSE++ \cite{faghri2017vse++}              &32.9 &61.6  &74.7   &24.1 &52.0 &66.2   & 51.9\\
\textbf{Ours}  &\bf{40.2} &\bf{70.1} &\bf{81.3}  &\bf{31.3} &\bf{61.5} &\bf{73.9}  & \bf{59.7}\\

\hline
VSE++ (Res) \cite{faghri2017vse++}      &41.3 &69.2  &81.2   &30.3 &59.1 &72.4  & 58.9\\
\textbf{Ours (Res)} &\bf{42.8} &\bf{72.3} &\bf{83.0} &\bf{33.1} &\bf{62.9} &\bf{75.5}  &\bf{61.6} \\

\hline
\hline
\end{tabular}
\label{table:coco5k}
\end{table}

\begin{figure*}[t]
\centering
\includegraphics[scale=0.47]{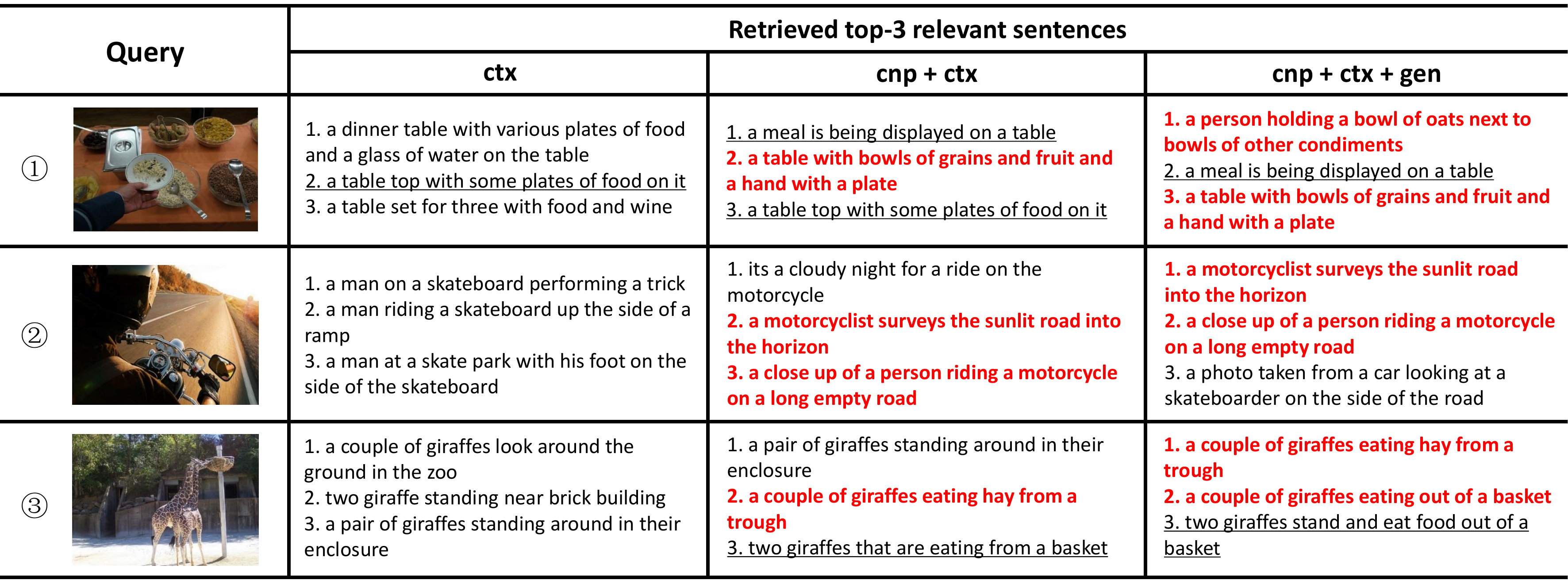}
\caption{Results of image annotation by 3 ablation models.
Groundtruth matched sentences are marked as red and bold, while some sentences sharing similar meanings as groundtruths are marked as underline (best viewed in colors).}
\label{fig:retrieval}
\end{figure*}

\subsection{Analysis of Image Annotation Results}

\begin{figure}[h]
\centering
\includegraphics[scale=0.54]{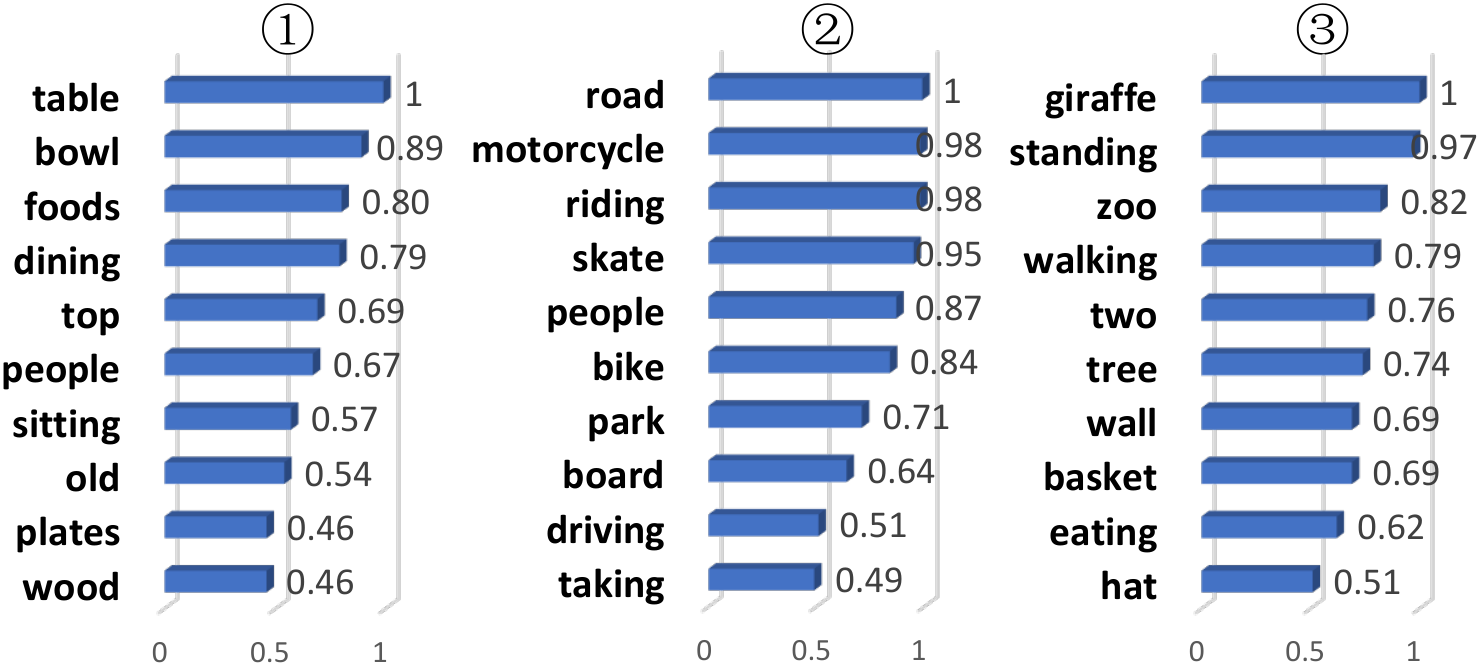}
\caption{Predicted top-10 semantic concepts with their confidence scores from the 3 images.}
\label{figure:concept}
\end{figure}

To qualitatively validate the effectiveness of our proposed model,
we analyze its image annotation results as follows.
We select several representative images with complex content,
and retrieve relevant sentences by 3 ablation models:
``ctx'', ``cnp + ctx'' and ``cnp + ctx + gen''.
We show the retrieved top-3 relevant sentences by the 3 models in Figure \ref{fig:retrieval},
and the predicted top-10 semantic concepts with confidence scores in Figure \ref{figure:concept}.

From Figure \ref{figure:concept}, we can see that our multi-regional multi-label CNN
can accurately predict the semantic concepts with high confidence scores
for describing the detailed image content. For example,
\emph{road}, \emph{motorcycle} and \emph{riding} are predicted from the second image.
We also note that the \emph{skate} is incorrectly assigned, which might result from the
reason that this image content is complicated and
the smooth country road looks like some skating scenes.

As shown in Figure \ref{fig:retrieval}, without the aid of the predicted semantic concepts,
``ctx'' cannot accurately capture the semantic concepts from complex image content.
For example, the retrieved sentences contain some clearly wrong semantic concepts
including \emph{water} and \emph{wine} for the first image,
and lose important concepts such as \emph{eating} and \emph{basket} for the third image.
After incorporating the predicted semantic concepts,
the retrieved sentences by ``cnp + ctx''
have very similar meanings as the images,
and are able to rank groundtruth sentences into top-3.
But the top-1 sentences still do not involve partial image details,
\emph{e.g.}, \emph{bowl}, \emph{sun} and \emph{eating} for the three images, respectively.
By further learning the semantic order\footnote{It is difficult to visualize the learned semantic order
like semantic concepts, since it is implied in the image representation.
Here we validate its effectiveness by showing it can help to find more accurate sentences.}
 with sentence generation,
the ``cnp + ctx + gen'' is able to associate all the related concepts
and retrieve the matched sentences with all the image details.

\section{Conclusions and Future Work}
In this work, we have proposed a semantic-enhanced image and sentence matching model.
Our main contribution is improving the image representation by
learning semantic concepts and then organizing them in a correct semantic order.
This is accomplished by a series of model components in
terms of multi-regional multi-label CNN, gated fusion unit,
and joint matching and generation learning.
We have systematically studied the impact of these
components on the image and sentence matching,
and demonstrated the effectiveness of our model by achieving
significant performance improvements.

In the future, we will replace the used VGGNet with ResNet in the multi-regional
multi-label CNN to predict the semantic concepts more accurately,
and jointly train it with the rest of our model in an end-to-end manner.
Our model can perform image and sentence matching and sentence
generation, so we would like to extend it for the image captioning task.
Although Pan \etal \cite{pan2016jointly} have shown the effectiveness
of using visual-semantic embedding for video captioning,
yet in the context of image captioning, its effectiveness has not been
well investigated.

{\small
\bibliographystyle{ieee}
\bibliography{egbib}
}

\end{document}